\documentclass{article} 
\usepackage{iclr2027_conference,times}

\usepackage{amsmath,amsfonts,bm}

\def\eqref#1{equation~\ref{#1}}

\def\1{\bm{1}}

\DeclareMathAlphabet{\mathsfit}{\encodingdefault}{\sfdefault}{m}{sl}
\SetMathAlphabet{\mathsfit}{bold}{\encodingdefault}{\sfdefault}{bx}{n}

\usepackage{hyperref}
\usepackage{url}
\usepackage{booktabs}
\usepackage{graphicx}
\usepackage{multirow} 
\usepackage[skins,breakable]{tcolorbox}
\title{Beyond On-Policy vs. Off-Policy: At What Level Should LLM Agent Distillation Follow the Student?}

\author{Ya Gao$^{1}$, Zhengtao Zou$^{1}$, Pekka Marttinen$^{1}$\\
$^1$Aalto University\\
\texttt{ya.gao@aalto.fi}
}
\iclrfinalcopy 
\begin{document}

\maketitle

\begin{abstract}
Distilling interactive LLM agents requires choosing between teacher-generated demonstrations and student-generated experience, corresponding broadly to off-policy and on-policy distillation. While both have proven effective, we argue that this distinction is inherently hierarchical: the policy that induces an interaction history need not be the same as the policy that generates the token prefix of the current action. We introduce a hierarchical agent distillation framework that formalizes these two levels of on-policiness and enables explicit control over their pairing, mixture proportions, and training schedule. We study this problem through self-distillation, where a frozen teacher has access to task guidance unavailable to the student, and conduct a three-stage controlled study across three models and four interactive environments. We find that student-induced histories paired with teacher-generated action prefixes generally provide the strongest standalone training signal, whereas following the student at both levels generally performs worst. Nevertheless, fully on-policy data becomes beneficial when incorporated into a mixture with other configurations. Moreover, mixtures with identical teacher/student proportions at each level can yield substantially different performance, showing that how the two levels are paired matters beyond the overall amount of on-policy data. Finally, front-loading fully teacher-generated contexts and delaying fully on-policy contexts improves over a static mixture. These findings provide practical principles for deciding where, how much, and when agent distillation should follow the student.

\end{abstract}

\section{Introduction}
\label{sec:introduction}

Knowledge distillation (KD)~\citep{hinton2015distilling} is widely used to post-train large language model (LLM) agents~\citep{yao2022react}, transferring task-solving behavior from a teacher to a student~\citep{zeng-etal-2024-agenttuning,kumar2026llm}. In \emph{privileged-information (PI) distillation}~\citep{penaloza2026privileged,alakuijala2026memento}, the teacher receives task guidance unavailable to the student, and distillation trains the student to reproduce the resulting behavior without that guidance at deployment.

A central design choice is which policy generates the contexts used for distillation. Off-policy approaches train on teacher-generated trajectories~\citep{kang2026distilling,liu2026structuredagentdistillationlarge}, providing coherent demonstrations but limited exposure to states encountered by the student. On-policy distillation (OPD)~\citep{agarwal2024policy,lu2025onpolicydistillation} instead obtains teacher supervision on student-generated experience, reducing training--deployment mismatch. However, student errors can produce histories or partial actions from which even the teacher has difficulty providing useful continuations~\citep{fu2026revisiting}. Both approaches can be effective in agent training~\citep{alakuijala2026memento,liu2026structuredagentdistillationlarge,xiao2026mimo,xu2026deepseek}, raising a basic question: \emph{should agent distillation be on-policy?}

We argue that the trajectory-level notion of on-policiness commonly used to characterize distillation methods is incomplete for interactive agents. An agent's training context contains two nested levels. At the \emph{history level}, preceding interactions determine the context and environment state before the current action. At the \emph{action-prefix level}, previously generated tokens within the current action determine the context for next-token prediction. Either level can independently come from the teacher or the student. For example, a teacher can generate an action from a student-induced history, while a student can generate an action from a teacher-induced history. The relevant question is therefore not only whether training is on-policy, but \emph{at which level and to what extent should distillation follow the student?}

We introduce \emph{hierarchical agent distillation}, which independently specifies the history source and action-prefix source. Choosing teacher or student at each level yields four configurations, $TT$, $ST$, $TS$, and $SS$, under a common distillation objective. This formulation separates three design choices: \emph{where} the student is followed, \emph{how much} each source configuration contributes to training, and \emph{when} different configurations are emphasized during learning.

We study these choices through a controlled three-stage investigation. Stage~1 isolates the effects of history and action-prefix sources and their interaction. Stage~2 examines how the four configurations should be paired and mixed under a fixed training budget. Stage~3 studies how their allocation should change over training. We use self-distillation, initializing teacher and student from the same checkpoint while providing task guidance only to the frozen teacher, thereby controlling for architecture, model size, and initialization. Across three backbones and four interactive environments, we obtain the following findings:

\begin{itemize}

\item \textbf{The value of on-policy training depends on the level.}
Student-induced histories paired with teacher-generated action prefixes ($ST$) generally provide the strongest standalone performance, while fully on-policy training ($SS$) performs worst across the tested settings. Student histories expose the teacher to more opportunities for corrective supervision, whereas student-generated prefixes can constrain the usefulness of teacher continuations.

\item \textbf{Pairing matters beyond the amount of on-policy data.}
Mixtures with identical proportions of student-generated histories and action prefixes can perform differently depending on how the two sources are paired. The strongest tested mixtures emphasize hybrid configurations while retaining contributions from both fully teacher-generated and fully on-policy contexts. In particular, a modest allocation to $SS$ can improve mixed training despite its weak standalone performance.

\item \textbf{The timing of data exposure matters.}
Front-loading $TT$ accelerates early learning, whereas delaying $SS$ improves later and final validation performance. Combining both schedules outperforms a static allocation throughout training and achieves the strongest final validation performance among the schedules tested.

\end{itemize}

\section{Hierarchical Agent Distillation}
\label{sec:framework}

\subsection{Preliminaries}
\label{sec:preliminaries}



An interactive LLM agent alternates between generating actions and receiving environment feedback. Let $h$ denote the interaction history before the current action, including the task instruction and preceding actions and observations. The action is generated autoregressively, and $u$ denotes a token prefix within it. The student predicts the next token according to
$\pi_\theta(\cdot\mid h,u)$.
The teacher additionally receives privileged task guidance $g$ and predicts
$\pi_T(\cdot\mid h,u,g)$.
Distillation trains the student to match this informed teacher while operating without $g$ at deployment.

The contexts $(h,u)$ used for distillation may be generated by either policy.
\emph{Teacher-trajectory context distillation} uses teacher-generated histories and action prefixes~\citep{alakuijala2026memento}, whereas \emph{fully on-policy distillation} uses student-generated histories and prefixes and queries the teacher for supervision on these contexts~\citep{agarwal2024policy,wang2026skill}.
We study self-distillation, initializing teacher and student from the same checkpoint and keeping the teacher frozen. The fully on-policy case therefore corresponds to on-policy self-distillation (OPSD)~\citep{zhao2026selfdistilled}.

\subsection{Hierarchical agent distillation framework}

We decompose on-policiness into two nested choices: which policy generates the interaction history, and which policy generates the current-action prefix conditioned on that history. This decomposition allows the two sources to be controlled independently, combined in different proportions, and scheduled over training.

\subsubsection{Two Levels of On-Policiness}
\label{sec:two_levels}

Let $H\in\{T,S\}$ denote the \emph{history source} and
$A\in\{T,S\}$ the \emph{action-prefix source}, where $T$ and $S$ denote the teacher and current student.
For a fixed data-collection round, let $d_H(h,g)$ denote the distribution of histories induced by source $H$, and let
$r_A(u\mid h,g)$ denote the distribution of action prefixes generated by source $A$ from history $h$.
Student generation ignores $g$.
The history source determines the preceding interaction trajectory, while the prefix source generates tokens only within the selected current action; the prefix therefore always remains conditioned on the chosen history.

\begin{figure*}[t]
    \centering
    \includegraphics[width=0.8\textwidth]{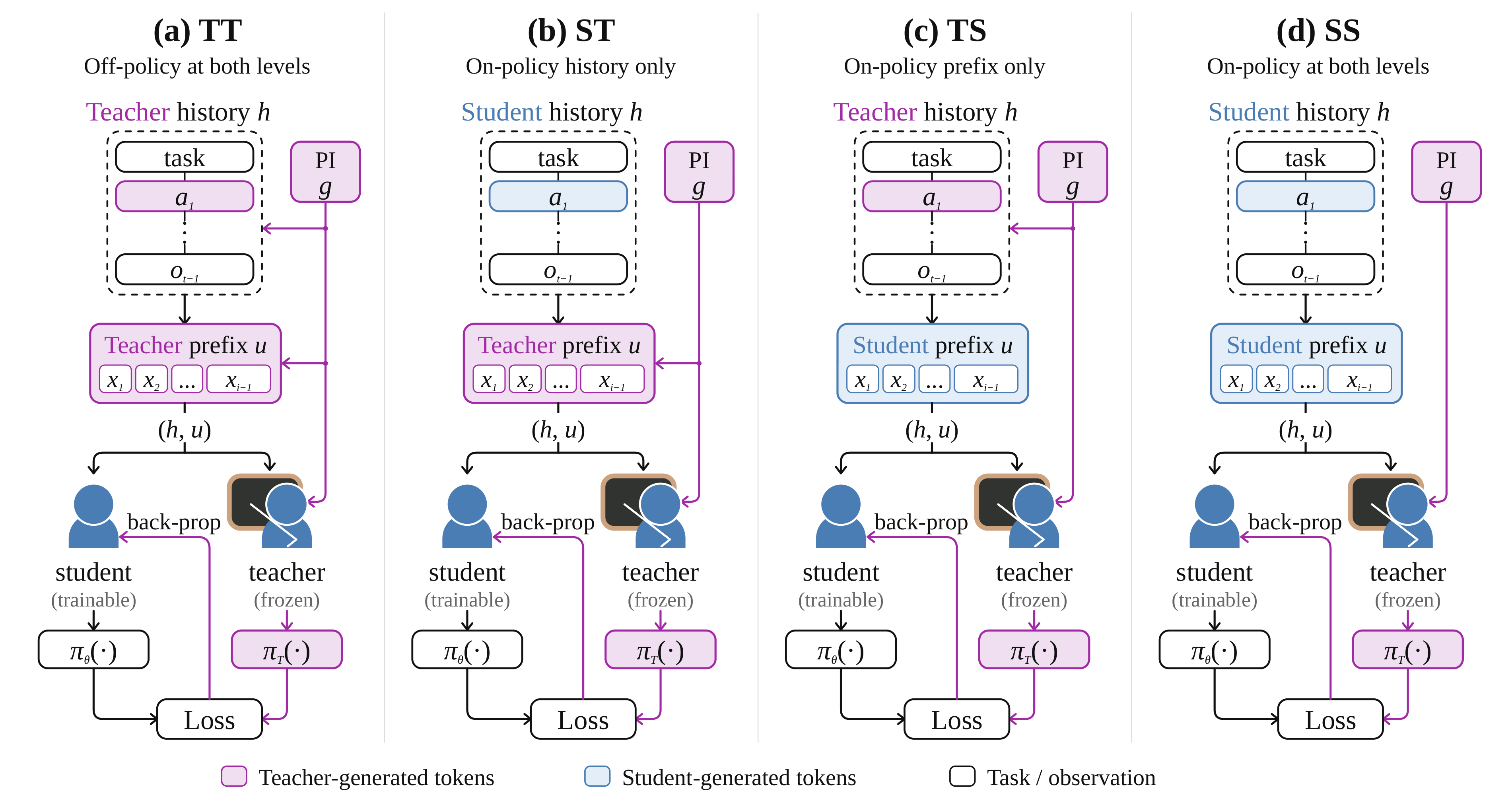}
    \caption{Illustration of the four source configurations.
The first letter denotes the interaction-history source and the second the current-action-prefix source.
In all configurations, the PI-conditioned teacher provides distillation supervision, while gradients update only the student.}

    \label{fig:framework}
\end{figure*}

Training is on-policy at the history level when $H=S$ and at the action-prefix level when $A=S$.
The four combinations are summarized in Figure~\ref{fig:framework}.
$TT$ corresponds to teacher-trajectory context distillation, while $SS$ recovers fully on-policy self-distillation.
$ST$ obtains teacher continuations at student-induced histories, analogous to expert supervision on learner-visited states in dataset aggregation~\citep{ross2011reduction}.
$TS$ instead generates student prefixes from teacher-induced histories, as in prefix replay.
In all four configurations, the PI-conditioned teacher provides the distillation target.


\subsubsection{A Common Training Objective}
\label{sec:training_objective}

For configuration $HA$, the student minimizes
\begin{equation}
\label{eq:source_loss}
\mathcal{L}_{HA}(\theta)
=
\mathbb{E}_{(h,g)\sim d_H}
\mathbb{E}_{u\sim r_A(\cdot\mid h,g)}
\left[
\mathcal{D}\!\left(
\pi_T(\cdot\mid h,u,g),
\pi_\theta(\cdot\mid h,u)
\right)
\right],
H\in\{T,S\}, 
A\in\{T,S\}.
\end{equation}
where $\mathcal{D}$ is the token-level distillation divergence. Forward KL uses $\mathcal{D}(p,q)=D_{\mathrm{KL}}(p\|q)$, and reverse KL uses $\mathcal{D}(p,q)=D_{\mathrm{KL}}(q\|p)$. The divergence and the two context sources are separate choices.

Operationally, we select a history from source $H$, generate the current action from source $A$, and evaluate teacher--student agreement along its token prefixes. The sampled context and teacher outputs are treated as fixed during each update; gradients are applied only to the student's next-token distribution. The same supervision rule therefore applies to all four configurations.




\subsubsection{Mixtures and Curricula}
\label{sec:mixtures_curricula}
Let $w_{HA}\geq 0$ denote the weight of configuration $HA$, with
$\sum_{H,A} w_{HA}=1$.
A training mixture minimizes
\begin{equation}
\label{eq:mixture_loss}
\mathcal{L}_{w}(\theta)
=
\sum_{H,A\in\{T,S\}}
w_{HA}\mathcal{L}_{HA}(\theta).
\end{equation}

The marginal weight on student histories is
$\alpha=w_{ST}+w_{SS}$,
and that on student action prefixes is
$\beta=w_{TS}+w_{SS}$.
Holding $\alpha$ and $\beta$ fixed while changing the joint weights changes only how the two source choices are paired.
For example, when $\alpha=\beta=\frac{1}{2}$,
equal weights on $TT$ and $SS$ form a \emph{diagonal} mixture,
equal weights on $ST$ and $TS$ form a \emph{crossed} mixture,
and equal weights on all four configurations form an \emph{independent} mixture.
These mixtures have identical marginal on-policiness but different source pairings.

Finally, allowing the weights to vary with training progress,
$w=w^{(k)}$,
defines a curriculum.
Schedules with matched cumulative configuration weights isolate \emph{when} each type of context is most useful.
The framework therefore separates three choices: \emph{where} to follow the student, \emph{how much} each source configuration contributes, and \emph{when} each should be emphasized.

\section{Setup of Controlled Study}
\subsection{Models and Benchmarks}
We use Qwen3-8B, Qwen3-14B~\citep{yang2025qwen3}, and Ministral-3-14B-Instruct-2512~\citep{liu2026ministral}, abbreviated as Ministral-3. For each backbone, teacher and student are initialized from the same checkpoint; the teacher remains frozen throughout training. We disable thinking mode for Qwen3.

We evaluate on ToolQA~\citep{zhuang2023toolqa}, OfficeBench~\citep{wang2024officebench}, ALFWorld~\citep{shridhar2021alfworld}, and WebShop~\citep{yao2022webshop}. Following \citet{alakuijala2026memento}, ToolQA and OfficeBench use train/validation/test splits of 646/107/323 and 191/9/86 tasks. For ALFWorld, we stratify by task type to sample 3,000 training and 200 validation tasks from the original training pool and retain the original evaluation split. For WebShop, we sample 3,000 training and 200 validation tasks and use the full 500-task test set. Further details are in Appendix~\ref{app:benchmarks}.

\paragraph{Privileged information construction.} 
The teacher receives task guidance describing general solution strategies, tool-use tips, and common failure modes. For ToolQA and OfficeBench, we initialize guidance from \citet{alakuijala2026memento} and refine it with Qwen3-32B using failed guidance-conditioned rollouts. For ALFWorld, we manually draft guidance per task type from failed student trajectories, use it to obtain successful Qwen3-32B trajectories, and let GPT-5.4 refine the guidance from both successes and failures. For WebShop, we manually draft guidance from 50 failed trajectories, organize it by decision stage and constraint pattern, and refine it with GPT-5.4 using corresponding successful Qwen3-32B trajectories. Examples are provided in Appendix~\ref{app:pi_construct}.

\subsection{Training Setup}
\paragraph{Agent architecture.}
Agents follow a ReAct-style interaction format~\citep{yao2022react}: at each step, the model generates a reasoning segment followed by an operation conditioned on the interaction history. Operations invoke tools, execute Python code, or issue environment-specific commands.

\paragraph{Data collection and refresh.}
Before training, we collect one trajectory from the initial student and cache three trajectories from the frozen teacher for each training and validation task. Student trajectories provide $SS$ examples; $ST$ retains their histories but regenerates the current action with the teacher. A randomly selected cached teacher trajectory provides $TT$ examples and the histories for $TS$, whose current actions are generated by the student. After each training round, we refresh all student-dependent data using the latest student: we regenerate student trajectories for $SS$, teacher continuations at those histories for $ST$, and student continuations at cached teacher histories for $TS$. $TT$ continues to sample from the three cached teacher trajectories.

\paragraph{Training exercises.}
Each exercise consists of an interaction history and one current action. We uniformly sample eight action positions per source trajectory, with replacement when necessary, so each task contributes the same number of exercises. Distillation conditions on the history and supervises tokens in the current action.

\paragraph{Optimization.} 
We apply rank-128 LoRA adapters to all linear layers. Training runs for eight rounds, refreshing student-dependent data between rounds. Each round contains 32, 16, 96, and 96 gradient updates for ToolQA, OfficeBench, ALFWorld, and WebShop, respectively, with global batch size 64. Full hyperparameters are in Appendix~\ref{app:hp}.

\subsection{Evaluation}
We evaluate each trained student on every test task with three trials. The maximum action counts for ToolQA, OfficeBench, ALFWorld, and WebShop are 15, 15, 32, and 15, respectively. We report mean success rate (SR), pass@3 (the fraction of tasks solved in at least one trial), and mean action count over all trajectories. For WebShop, we additionally report the official Task Score, averaged across trials, which gives partial credit for satisfying task requirements. Decoding settings are listed in Table~\ref{tab:eval_decoding}.

\section{Stage 1: Separating the Two Levels of On-Policiness}
\label{sec:stage1}
\begin{table*}[t]
\centering

\caption{
Stage 1 results with forward KL. Higher SR, pass@3, and Task Score is better. Bold and underlined values mark the best and second-best performance among the four distilled variants of each backbone. Subscripts denote standard error over three trials.}
\label{tab:stage1_main_results}
\resizebox{0.99\textwidth}{!}{%
\begin{tabular}{
l
ccc
ccc
ccc
cccc
}
\toprule
\multirow{2}{*}{\textbf{Model}}
& \multicolumn{3}{c}{\textbf{ToolQA}}
& \multicolumn{3}{c}{\textbf{OfficeBench}}
& \multicolumn{3}{c}{\textbf{ALFWorld}}
& \multicolumn{4}{c}{\textbf{WebShop}} \\
\cmidrule(lr){2-4}
\cmidrule(lr){5-7}
\cmidrule(lr){8-10}
\cmidrule(lr){11-14}

& \textbf{SR}
& \textbf{pass@3}
& \textbf{Action Count}
& \textbf{SR}
& \textbf{pass@3}
& \textbf{Action Count}
& \textbf{SR}
& \textbf{pass@3}
& \textbf{Action Count}
& \textbf{SR}
& \textbf{pass@3}
& \textbf{Task Score}
& \textbf{Action Count} \\
\midrule

Qwen3-8B
&17.2$_{2.7}$ &20.1 &11.3& 7.0$_{0.9}$ & 8.1&14.2 & 37.2$_{1.7}$&48.2 & 30.1& 6.6$_{1.0}$&9.4 &38.1$_{1.5}$ &11.3 \\
\quad + TT
&\textbf{81.7}$_{0.9}$ & \underline{89.2}&6.3 & 70.9$_{3.3}$& 76.7&9.1 & \underline{69.7}$_{2.3}$& \underline{77.4}&16.7 &\underline{70.2}$_{0.7}$ &78.0 &\underline{78.5}$_{0.8}$ & 7.9\\
\quad + ST
&\underline{81.1}$_{1.5}$ &\textbf{91.3} &6.8 & \textbf{74.8}$_{2.1}$&\textbf{84.9} & 8.2&\textbf{71.7}$_{2.0}$ &\textbf{79.9} & 15.1&\textbf{73.3}$_{0.2}$ &\textbf{83.2} & \textbf{82.2}$_{1.0}$& 7.1\\
\quad + TS
&79.8$_{1.1}$&87.9 &6.4 &\underline{71.7}$_{2.5}$ &\underline{80.2} & 8.7& 69.3$_{1.5}$&75.5 & 15.4&69.1$_{0.9}$ &\underline{79.2} &75.0$_{0.3}$ & 7.5\\
\quad + SS
&74.5$_{1.7}$&82.4 &7.5 & 65.9$_{1.8}$& 73.3& 9.6& 60.8$_{1.8}$& 67.9&20.0&58.9$_{1.1}$ &66.8 &73.1$_{0.9}$ & 9.9\\

\midrule

Qwen3-14B
&19.8$_{2.8}$ &24.1 &11.7 & 7.8$_{0.4}$&10.5 &13.1&42.3$_{2.6}$ & 52.9&26.9 & 9.1$_{0.7}$ &13.8 &50.6$_{0.7}$ &12.5 \\
\quad + TT
& \underline{84.9}$_{1.2}$&\underline{91.3} &6.0 &\underline{75.2}$_{0.9}$ & \underline{84.9}& 8.3&72.1$_{1.9}$ & \underline{80.7}& 16.2&75.6$_{0.8}$ &81.2 &\underline{82.7}$_{0.5}$ & 7.4\\
\quad + ST
& \textbf{85.2}$_{1.4}$&\textbf{92.6} & 6.2& \textbf{77.5}$_{1.4}$& \textbf{89.5}&7.6 &\textbf{74.5}$_{1.4}$ &\textbf{83.9} & 15.3&\textbf{79.4}$_{1.3}$ & \textbf{86.8}&\textbf{85.0}$_{0.6}$ & 7.2\\
\quad + TS
&84.3$_{0.7}$ &88.9 & 6.9& 74.8$_{1.3}$&82.6 &8.5 & \underline{72.6}$_{2.0}$&80.3 & 15.9&\underline{76.0}$_{1.2}$ & \underline{83.0}&81.9$_{1.2}$& 7.7 \\
\quad + SS
&79.1$_{1.2}$ & 85.1&7.1 & 70.9$_{0.8}$& 79.1& 8.9&67.8$_{2.3}$ &76.6 & 19.4&67.9$_{0.9}$ &77.0 & 78.8$_{0.4}$& 8.5\\

\midrule

Ministral-3
& 15.4$_{1.9}$ & 21.1&10.0 &$8.5_{1.3}$ & 10.5&12.0 &38.4$_{1.6}$ &49.6 &29.1 &$9.7_{0.8}$ &12.6 &38.9$_{1.2}$ &10.8 \\
\quad + TT
& \textbf{82.4}$_{0.9}$&\underline{90.1} &5.8 &73.3$_{0.6}$ & 81.4& 8.7&71.0$_{1.3}$ &\underline{79.9} & 17.2&\underline{72.6}$_{0.4}$ &79.6 &\underline{79.9}$_{1.1}$ &8.0 \\
\quad + ST
& \underline{81.8}$_{1.3}$&\textbf{93.2} & 5.5&\textbf{76.0}$_{1.0}$ &\textbf{86.0} & 8.1&\textbf{74.3}$_{1.0}$ & \textbf{81.8}& 15.8&\textbf{75.5}$_{1.0}$ &\textbf{83.4} &\textbf{80.8}$_{1.0}$ & 7.6\\
\quad + TS
& 80.9$_{0.8}$& 85.4& 5.1&\underline{73.6}$_{1.8}$ &\underline{82.6} &8.5 &\underline{71.5}$_{1.1}$ & 79.6& 17.7&71.9$_{1.0}$ &\underline{80.0} &77.2$_{0.8}$ & 7.8\\
\quad + SS
& 78.2$_{1.1}$&84.8 &6.4 &71.3$_{0.9}$ & 79.1&9.2 &66.5$_{1.6}$ &75.9 & 18.6&64.4$_{1.1}$ &74.4 &73.4$_{0.6}$ & 8.4\\

\bottomrule
\end{tabular}
}
\end{table*}

To isolate the effects of history- and action-prefix-level on-policiness, we train one student per configuration, matching initialization, training tasks, and the number of exercises per task within each backbone--benchmark setting.

\subsection{Main Results}
\label{sec:stage1_main}
Table~\ref{tab:stage1_main_results} reports results with forward KL. All four configurations substantially improve over the initial student, but their relative performance reveals a consistent distinction between the two levels.

\paragraph{The effect of student histories depends on the action-prefix source.}
Replacing teacher histories with student histories generally improves success rate when prefixes are teacher-generated ($TT \rightarrow ST$), but consistently hurts when prefixes are student-generated ($TS \rightarrow SS$). Likewise, teacher prefixes provide little advantage on teacher histories ($TT$ vs.\ $TS$: at most 1.9 points), but substantially outperform student prefixes on student histories ($ST$ vs.\ $SS$: 3.6--14.4 points). Thus, student-induced histories are most effective when paired with teacher-generated actions.

\paragraph{$ST$ performs best overall, while $SS$ is weakest standalone.}
$ST$ achieves the highest success rate in 10 of 12 settings, the highest pass@3 in all 12, and the highest WebShop Task Score for all three backbones. The two exceptions are ToolQA, where $TT$ exceeds $ST$ by 0.6 points for Qwen3-8B and Ministral-3, although $ST$ still attains higher pass@3.
In contrast, $SS$ has the lowest success rate and pass@3 among distilled variants in every setting. Its average success-rate deficit relative to $ST$ is larger for Qwen3-8B than Qwen3-14B (10.2 vs.\ 7.7 points), indicating greater sensitivity of the smaller model to student-generated prefixes.

\paragraph{The pattern is robust to KL direction.}
Reverse-KL results in Appendix~\ref{app:reverse_kl} preserve the main source-level trends. Forward KL is generally similar or better, except in some $SS$ settings, and is used throughout the remaining experiments.

\subsection{Analysis}
\label{sec:stage1_analysis}

We next examine why the source choices interact and whether configurations that are weaker standalone nevertheless learn complementary behaviors. Annotation protocols are described in Appendix~\ref{app:traj_analysis}.

\subsubsection{Supervision during data collection}

\paragraph{Student histories provide richer corrective supervision ($TT$ vs.\ $ST$).}
For each benchmark, we sample 100 training tasks and two exercises per task. We measure \emph{Unresolved}, the fraction of histories containing an uncorrected earlier error; \emph{Identified}, the fraction of unresolved cases where the current action recognizes the error; and \emph{Addressed}, the fraction of identified cases receiving a resolution or appropriate corrective step.
Unresolved errors occur in 82.7--85.0\% of $ST$ exercises but only 14.0--19.1\% of $TT$ exercises (Table~\ref{tab:stage_1_tt_vs_st}). Although identification is lower under $ST$, identified errors are addressed at comparable rates, giving $ST$ substantially more opportunities for corrective supervision. Across training rounds, unresolved errors decrease while the teacher identifies a larger fraction of those remaining (Figure~\ref{fig:stage1_tt_vs_st}).

\paragraph{Student prefixes constrain teacher continuations ($ST$ vs.\ $SS$).}
We select 100 student-history exercises with unresolved errors per benchmark and compare a fully teacher-generated action with teacher continuations conditioned on the student's reasoning, or on its reasoning plus half of its operation. We measure continuation \emph{Executability} and \emph{Appropriateness}.
Both decline as more student-generated prefix is retained (Table~\ref{tab:stage_1_ss_vs_st}); with half the operation retained, appropriateness drops from 80.2\% to 56.2\% for Qwen3-8B and from 83.8\% to 60.2\% for Qwen3-14B. Thus, student prefixes can constrain the teacher into less useful continuations even at the same history.

\paragraph{History sources change where teacher and student disagree ($TS$ vs.\ $SS$).}
At rounds 1, 3, 5, and 8, we sample 100 exercises per configuration containing an explicit error without subsequent self-correction (see details in Appendix~\ref{app:traj_analysis}). We compute mean per-token KL before, within, and after the first annotated error span, then normalize the three regional means by their sum. Under $TS$, disagreement concentrates on the error span, whereas $SS$ assigns relatively more disagreement before the error (Figure~\ref{fig:stage1_ss_vs_ts}). Teacher histories therefore produce more error-localized disagreement, while student histories distribute disagreement more broadly across the action.

\subsubsection{Behavior of trained students}

\paragraph{Configurations learn different behavioral strengths.}

For 80 test tasks per benchmark, we annotate one trajectory per configuration using five nonexclusive error categories: repeated failure, premature completion, rule inconsistency, operational mistakes despite correct plans, and high-level planning mistakes.
$SS$ exhibits fewer repeated failures and planning mistakes, whereas $TT$ follows explicit rules more reliably (Table~\ref{tab:stage_1_mistake_type}). 

\paragraph{$ST$ is strongest at recovering from student errors.}

For Qwen3-14B, we resume all four trained policies from the same 100 erroneous initial-model histories per benchmark on ToolQA, ALFWorld, and WebShop. $ST$ achieves the highest rates of both error correction and eventual task success (Figure~\ref{fig:stage1_recovery}).
Although $TT$ and $TS$ outperform $SS$ from task starts, their eventual success from erroneous histories is similar to $SS$. Their standard-evaluation advantage therefore does not translate into stronger recovery. $ST$ uniquely combines exposure to student mistakes with teacher-generated corrective actions.

\paragraph{Configurations also differ in learning dynamics.}

For Qwen3-14B on ToolQA, ALFWorld, and WebShop, $TT$ leads during rounds 1--3, $ST$ overtakes it by round 5 and finishes highest, while $SS$ improves more gradually and continues gaining later in training (Figure~\ref{fig:stage1_validation}). These complementary behaviors motivate mixture design in Stage~2 and scheduling in Stage~3.

\section{Stage 2: Pairing and training-mixture study}
\label{sec:stage2}

Stage~1 favors $ST$ as a standalone configuration, but weaker configurations may provide complementary supervision when mixed. We therefore study how the four configurations should be paired and allocated under a fixed training budget. All mixtures match the total number of exercises and training rounds and reuse the initial Stage~1 exercise pools. We order mixture weights as $(TT,ST,TS,SS)$.

\subsection{Pairing Matters at Fixed On-Policiness}
\label{sec:stage2_pairing}

We compare three mixtures with identical marginal on-policiness:
$w_{\mathrm{Diagonal}}=(0.5,0,0,0.5)$,
$w_{\mathrm{Crossed}}=(0,0.5,0.5,0)$, and
$w_{\mathrm{Independent}}=(0.25,0.25,0.25,0.25)$.
Each assigns 50\% of examples to student histories and 50\% to student action prefixes, differing only in how the two sources are paired.

Figure~\ref{fig:stage2_main} shows Qwen3-14B results; other backbones are in Appendix~\ref{app:stage2_main}. The independent mixture performs best across all shown metrics and benchmarks and improves success rate over standalone $ST$ by 0.5--3.3 points. Thus, marginal on-policiness alone is insufficient: the joint pairing of history and prefix sources also matters.

\begin{figure*}[t]
    \centering
    \includegraphics[width=0.95\textwidth]{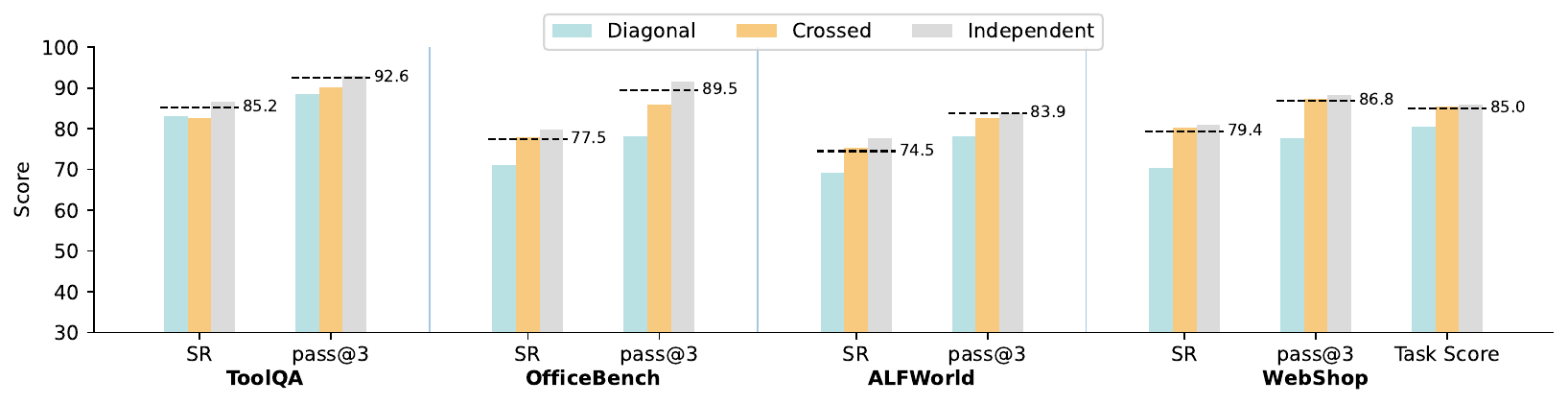}
    \caption{Pairing at fixed marginal on-policiness with Qwen3-14B. Dashed lines indicate standalone $ST$ performance from Stage~1.}
    \label{fig:stage2_main}
\end{figure*}

\subsection{From Complementarity to Mixture Composition}
\label{sec:stage2_composition}

\begin{figure*}[t]
    \centering
    \includegraphics[width=0.5\textwidth]{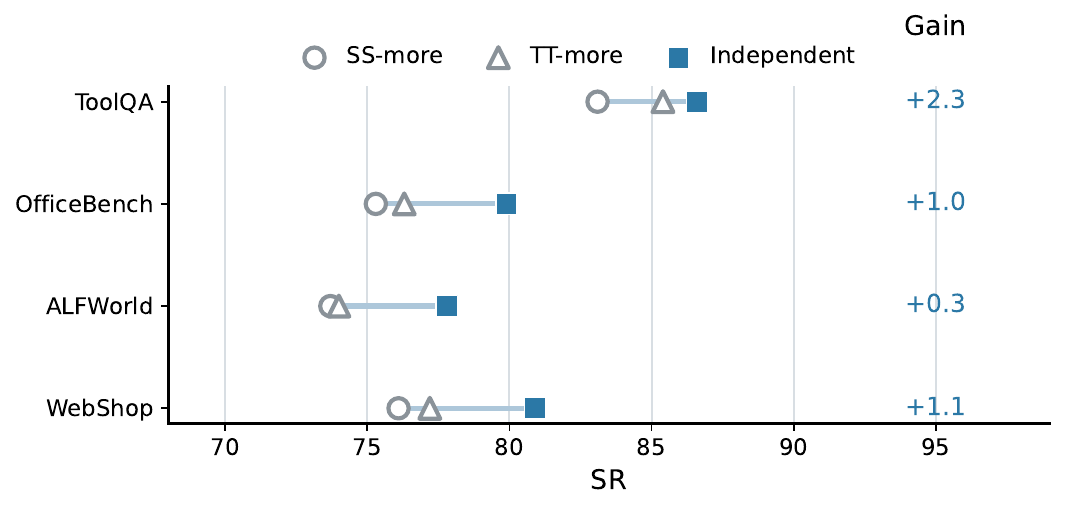}
    \caption{Retaining both diagonal configurations improves success rate with Qwen3-14B. Holding $ST$ and $TS$ at 0.25 each, the remaining allocation is assigned to $TT$ only (TT-more), $SS$ only (SS-more), or equally to both (Independent)}
    \label{fig:stage2_tt_vs_ss}
\end{figure*}

\paragraph{Retain both $TT$ and $SS$.}
Holding $ST$ and $TS$ at 0.25 each, we compare \emph{TT-more}, $(0.5,0.25,0.25,0)$, and \emph{SS-more}, $(0,0.25,0.25,0.5)$, against the independent mixture. Across all settings, the independent mixture outperforms both, while TT-more outperforms SS-more (Figures~\ref{fig:stage2_tt_vs_ss}, \ref{fig:stage2_tt_vs_ss_qwen3-8b}, and \ref{fig:stage2_tt_vs_ss_ministral}). Thus, retaining both diagonal configurations is more effective than assigning their combined budget entirely to either one. In particular, adding some $SS$ improves over TT-more despite $SS$'s weak standalone performance. The TT-more--SS-more gap is smaller for Qwen3-14B than for Qwen3-8B, consistent with the smaller model's greater sensitivity to fully on-policy training.

\paragraph{Emphasize hybrid configurations without removing the diagonals.}
We increase the combined weight of $ST$ and $TS$ from 0.5 to 0.75, yielding the \emph{hybrid-heavy} mixture $(0.125,0.375,0.375,0.125)$. It improves success rate over the independent mixture in every setting (Table~\ref{tab:stage_2_more_st_ts}). However, removing $TT$ and $SS$ entirely, as in the crossed mixture, performs worse than the independent mixture. These results favor emphasizing hybrid configurations while retaining complementary $TT$ and $SS$ examples.

\paragraph{Favor student histories over student action prefixes.}
Finally, we ask how to divide the hybrid allocation. Keeping $TT$ and $SS$ at 0.125 each, we compare \emph{ST-more}, $(0.125,0.5,0.25,0.125)$, against \emph{TS-more}, $(0.125,0.25,0.5,0.125)$. These mixtures exchange the student-history and student-prefix proportions, while preserving the combined hybrid allocation. ST-more yields higher success rate in every setting (Figure~\ref{fig:stage2_st_vs_ts}), extending the directional preference observed from Stage~1 to this mixed training.

The resulting principle of these comparisons is to emphasize student histories paired with teacher actions without excluding configurations that are weaker in isolation.

\section{Stage 3: Scheduling the Training Mixture}
\label{sec:stage3}
Stage~2 shows that mixture composition matters, but should the same allocation be maintained throughout learning? The dynamics analysis in Stage~1 suggest otherwise: $TT$ supports rapid initial progress, whereas $SS$ improves more slowly and continues gaining later. We therefore investigate whether emphasizing these configurations at different stages improves mixed training.

\begin{figure*}[t]
    \centering
    \includegraphics[width=0.95\textwidth]{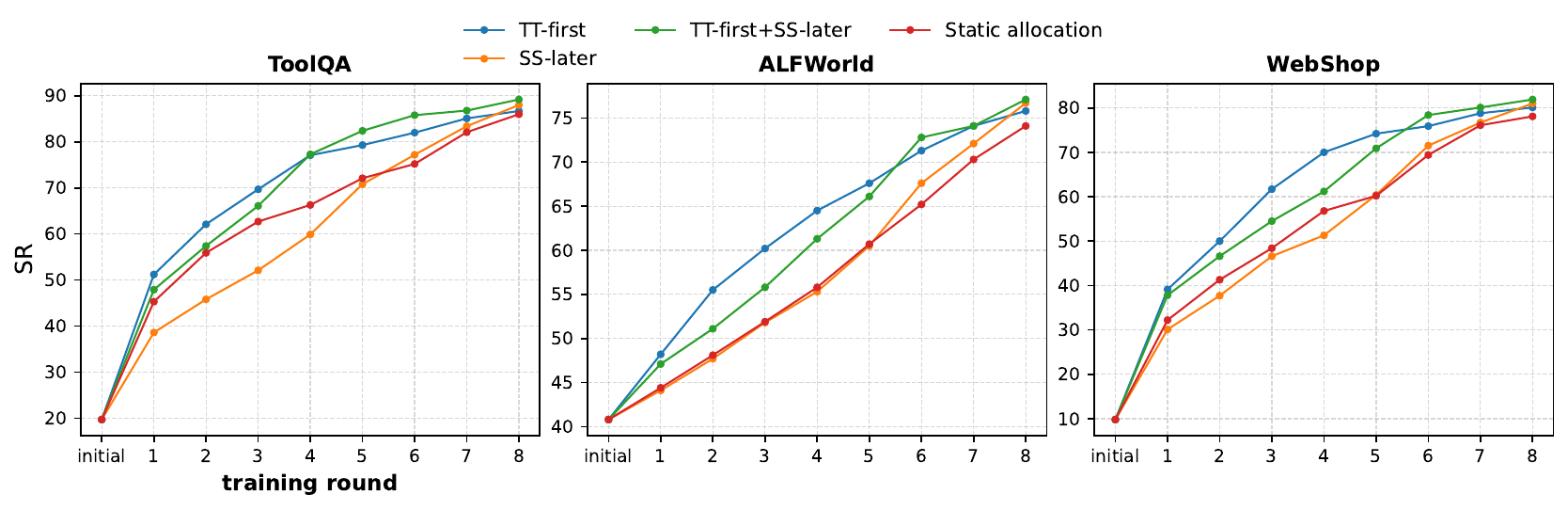}
    \caption{Validation success rate across eight training rounds under static and scheduled allocations of the independent mixture. \textit{initial} denotes the initial student.}
    \label{fig:stage3_validation}
\end{figure*}

We use the independent mixture, $(0.25,0.25,0.25,0.25)$, as a balanced reference and compare its static allocation with three schedules over 8 training rounds. \emph{TT-first} places approximately 80\% of the total $TT$ allocation in rounds 1--4 and the remaining 20\% in rounds 5--8. \emph{SS-later} reverses this split for $SS$, allocating 20\% early and 80\% late. \emph{TT-first + SS-later} combines both schedules. Each targeted configuration retains its total allocation across training. We measure validation success rate after every round on ToolQA, ALFWorld, and WebShop (Figure~\ref{fig:stage3_validation}).

$TT$-first accelerates initial learning on all three benchmarks: by round~4, its success rate exceeds the static baseline by 8.7--13.2 percentage points. Its final advantage is smaller, at 0.7--2.0 points. In contrast, $SS$-later initially trails the static baseline on ToolQA and WebShop and closely tracks it on ALFWorld, but overtakes it during the second half of training and finishes 2.0--2.9 points higher. The two schedules therefore offer distinct benefits: front-loading $TT$ improves early progress, while delaying $SS$ yields a larger improvement in final performance.

The combined schedule outperforms the static baseline at all evaluated training rounds and achieves the highest final validation success rate among the tested schedules on all three benchmarks. Although $TT$-first alone often leads early, the combined schedule ultimately surpasses it. These results suggest complementary temporal roles: teacher-generated contexts support rapid initial learning, while greater exposure to fully on-policy contexts is more beneficial later. They favor front-loading $TT$ and delaying $SS$ within a mixture, rather than maintaining a uniform allocation throughout training.

\paragraph{Conclusions across the three stages.}

Across the three stages, effective on-policiness depends on \emph{where}, \emph{how much}, and \emph{when} student-generated context is used. Student histories paired with teacher actions provide the strongest standalone performance and recovery, while mixed training benefits from retaining configurations that are weaker in isolation. Scheduling $TT$ earlier and $SS$ later further improves over a static mixture. Thus, effective agent distillation depends not on maximizing on-policy data, but on how it is sourced, combined, and scheduled.

\section{Related Work}
\label{sec:related_work}

\paragraph{Distillation and privileged information for agents.}
Knowledge distillation transfers knowledge from a teacher model to a student model~\citep{hinton2015distilling}. Demonstration-based approaches improve LLM agents through supervised training on interaction trajectories~\citep{zeng-etal-2024-agenttuning}, while context distillation transfers behaviors elicited by additional instructions or reasoning context into a model that no longer requires such context at inference time~\citep{snell2022learning, gao2026edit}. For interactive agents, \citet{alakuijala2026memento} internalize task guidance and corrective feedback through iterative distillation, whereas \citet{wang2026skill} use trajectory-derived skills as privileged teacher context for on-policy self-distillation. Our work addresses the complementary question of how teacher- and student-generated contexts should be constructed, combined, and scheduled when distilling privileged guidance.

\paragraph{On-policy distillation and imitation learning.}
On-policy distillation (OPD)~\citep{agarwal2024policy} mitigates training--deployment mismatch by obtaining teacher supervision on student-generated sequences, and has become increasingly common in LLM post-training~\citep{lu2025onpolicydistillation, zhao2026selfdistilled}. For multi-turn agents, however, student-generated interactions may reach states for which the teacher cannot provide reliable supervision. Concurrent work, ReOPD~\citep{liao2026multi}, addresses this issue by generating student actions from replayed teacher histories, enabling teacher supervision without requiring new environment interactions during student training. Related challenges arise in imitation learning~\citep{ross2011reduction}, where expert supervision is collected on learner-visited states to mitigate sequential distribution shift. Our framework distinguishes history-level and action-prefix-level on-policiness as separate but coupled design choices, thereby unifying teacher-generated, student-generated, and hybrid contexts under a common objective. This formulation enables controlled analysis of their individual effects, their interactions in mixed training regimes, and their allocation over the course of training.

\section{Conclusion}
\label{sec:conclusion}
We introduced hierarchical agent distillation framework that formalizes on-policiness at the history and action-prefix levels. Across three backbones and four interactive environments, student-induced histories paired with teacher-generated actions generally provide the strongest standalone performance. Mixture comparisons show that pairing matters even when marginal on-policiness is fixed, and that configurations weaker in isolation can improve mixed training. Scheduling experiments further show that emphasizing fully teacher-generated contexts early and fully on-policy contexts later improves validation performance over a static uniform mixture. Together, these findings move agent distillation beyond a binary on-policy/off-policy choice: effective training depends on where the student is followed, how teacher and student contexts are paired, and when each configuration is emphasized.
\paragraph{Limitation.} Our experiments focus on privileged-information-based self-distillation with frozen teachers initialized from the same checkpoints as the students. Generalization to substantially larger models remains to be established. Also, the mixture study identifies composition principles through targeted comparisons rather than a universally optimal allocation.

\subsection*{AI use statement}
We used large language models to assist with manuscript writing and editing, as well as in parts of the experimental pipeline, including annotating agent trajectories and constructing task guidance. AI-assisted outputs used in the experiments were generated following the procedures described in the paper and appendix and were reviewed or validated as appropriate. The authors remain responsible for the study design, experimental analysis, interpretation of results, and final manuscript.


\bibliography{iclr2027_conference}
\bibliographystyle{iclr2027_conference}

\appendix
\section{Appendix}
\subsection{Additional details on experiment setup}
\subsubsection{Privileged information examples}
\label{app:pi_construct}
We show the prompt used in constructing privileged information and PI examples.

\begin{tcolorbox}[breakable, colback=gray!5, colframe=gray!40, title=Prompt]
You are an expert at improving guidance for an LLM agent that must complete interactive tasks.

Your goal is to **optimize a set of seed hints/guidance** using examples of successful and failed agent trajectories. The optimized hints will later be provided to another agent to help it solve new tasks of the same kind.

Inputs: Task / Task Type

{TASK DESCRIPTION}

Seed Hints: {SEED HINTS}

Successful Trajectories: {SUCCESSFUL TRAJECTORIES}

Failed Trajectories: {FAILED TRAJECTORIES}

Your Task

Analyze the successful and failed trajectories and revise the seed hints into a better set of guidance for solving this type of task.

Use the trajectories **contrastively**:

- Identify strategies, decisions, action sequences, or checks that consistently contribute to success.

- Identify mistakes, inefficient behaviors, incorrect assumptions, or missing checks that contribute to failure.

- Pay particular attention to situations where successful and failed trajectories encounter similar states but take different actions.

- Determine whether any existing seed hint is incomplete, ambiguous, misleading, overly specific, or unnecessary.

- Preserve useful information from the seed hints while correcting or extending it when supported by the trajectories.

The resulting guidance should help an agent make better decisions during execution, rather than merely describe what happened in the demonstrations.

Requirements for the Optimized Hints

**Generalize beyond the demonstrations.** 

   Extract reusable strategies that apply to new tasks of the same type. Do not memorize task-specific object names, locations, answers, or exact action sequences unless they represent a general rule.

**Be actionable.**  

   State what the agent should do, when it should do it, and, when useful, what observations or conditions it should check before acting.

**Prioritize failure prevention.**  

   Explicitly address recurring mistakes observed in failed trajectories when they can be prevented by a general rule.

**Be concise.**  
   Include only information that is likely to improve task success. Merge redundant rules and remove low-value commentary.

Reasoning Procedure

Before producing the final hints, internally:

1. Compare successful and failed trajectories.

2. Identify important decision points and recurring failure modes.

3. Determine which seed hints already address them.

4. Add, remove, merge, or revise hints as necessary.

5. Check that every final hint is both useful and sufficiently general.

Output

Return only the optimized guidance in the following format:

\[ ...... \]

Do not summarize the trajectories, mention individual demonstrations, or explain how you derived the hints. The final output should be directly usable as guidance for an agent solving a new task.
\end{tcolorbox}

\begin{tcolorbox}[breakable, colback=gray!5, colframe=gray!40, title=Guidance Example]

Steps to Answer the Question:

1. Read the question carefully.

2. Choose appropriate tools. Read the documentation of the chosen tools to understand the functions of them.

3. Your final answer should be a string only containing the answers to the question without explanatory text.

Specific hint:

To solve the question asking about the total number of papers authored by A plus those written by the collaborators of A, you should follow:

1. Identify the co-author(s) of A. (Answer: B).

2. For A and B, get their joint paper by checking the edge attributes.

3. Find out additional co-authors of B.

4. Retrieve the papers jointly authored by B and each of B's additional co-authors.

5. Combine papers and remove duplicates to determine the total count
\end{tcolorbox}

\begin{tcolorbox}[breakable, colback=gray!5, colframe=gray!40, title=Guidance Example]

This is a heat-then-place task.

Required subgoals:

1. Locate and acquire the requested object.

2. Heat that object using a microwave.

3. Preserve the identity of the heated object.

4. Deliver the heated object to the requested receptacle.

Usefule action patterns:
\[......\]

Reasoning guidance:

- The final placement alone is insufficient: the object must first satisfy the ``hot" state requirement.

- Search observations for the exact target object.

- Once a target instance has been picked up, continue manipulating that same instance.

- Do not search for the destination until necessary if the target object has not yet been acquired.

\end{tcolorbox}

\subsubsection{Benchmarks}
\label{app:benchmarks}

We evaluate on four interactive agent benchmarks spanning tool-augmented question answering, office automation, embodied household tasks, and web interaction.

\begin{itemize}
    \item \textbf{ToolQA.}
    ToolQA~\citep{zhuang2023toolqa} evaluates an agent's ability to answer questions by retrieving and reasoning over information from external data sources with heterogeneous formats, including text documents, graphs, and tabular databases. Following the processed version of \citet{alakuijala2026memento}, we consider six task groups: \textit{DBLP}, \textit{Agenda}, \textit{Yelp}, \textit{Airbnb}, \textit{Flight}, and \textit{Coffee}. The agent has access to nine tools for querying and processing the corresponding data sources, while individual task groups typically require only a subset of these tools.

    \item \textbf{OfficeBench.}
    OfficeBench~\citep{wang2024officebench} evaluates an agent's ability to complete multi-step office workflows involving six applications: \textit{Email}, \textit{Calendar}, \textit{Word}, \textit{Excel}, \textit{OCR}, and \textit{PDF Reader}. The agent has access to 20 tools for interacting with these applications. Tasks can involve one to three applications, requiring the agent to plan across multiple steps, select appropriate tools, and transfer information between applications.

    \item \textbf{ALFWorld.}
    ALFWorld~\citep{shridhar2021alfworld} is a text-based embodied environment in which an agent follows natural-language instructions to perform household tasks through sequential interaction with objects and receptacles. We consider its six commonly used task categories: \textit{Pick \& Place}, \textit{Pick Two \& Place}, \textit{Clean \& Place}, \textit{Heat \& Place}, \textit{Cool \& Place}, and \textit{Examine in Light}. Completing a task requires the agent to explore the environment, locate relevant objects, and execute an appropriate sequence of navigation and manipulation actions, such as picking up, placing, cleaning, heating, or cooling objects.

    \item \textbf{WebShop.}
    WebShop~\citep{yao2022webshop} is a simulated e-commerce environment that evaluates agents on interactive web navigation and product search. Given a natural-language shopping instruction specifying desired product properties and constraints, the agent must search for candidate products, navigate between search and product pages, inspect product information and available options, and purchase an item that best satisfies the instruction. The environment exposes two main types of actions, \texttt{search[query]} and \texttt{click[option]}, requiring the agent to perform query formulation, information gathering, comparison, and multi-step decision making.
\end{itemize}
\subsubsection{Training data statistics and hyperparameter settings}
\label{app:hp}
We provide the training data statistics for each benchmark in Table~\ref{tab:training_statistics}.

We run all experiments on 8 H100 80 GB GPUs and 4 AMD MI250X GPUs. We use the Adam optimizer throughout all training. Hyperparameter settings are presented in Table~\ref{tab:training_hyperparameters}.

\begin{table*}[t]
\centering
\small
\setlength{\tabcolsep}{5pt}
\caption{Training data statistics}
\label{tab:training_statistics}
\begin{tabular}{lcccc}
\toprule
Parameter & ToolQA & OfficeBench & ALFWorld & WebShop \\
\midrule
Training tasks & 646 & 191 & 3,000 & 3,000 \\
Training rounds & 8 & 8 & 8 & 8 \\
Task visits per round & 256 & 128 & 768 & 768 \\
Total task visits & 2,048 & 1,024 & 6,144 & 6,144 \\
Equivalent task passes & 3.17 & 5.36 & 2.05 & 2.05 \\
Action examples per task visit & 8 & 8 & 8 & 8 \\
Global batch size & 64 & 64 & 64 & 64 \\
Optimizer updates per round & 32 & 16 & 96 & 96 \\
Total optimizer updates & 256 & 128 & 768 & 768 \\
\bottomrule
\end{tabular}

\end{table*}

\begin{table*}[t]
\centering
\small
\setlength{\tabcolsep}{5pt}
\caption{Training hyperparameters for the controlled study. All settings are shared across source configurations.}
\label{tab:training_hyperparameters}
\begin{tabular}{lcc}
\toprule
Parameter & Qwen3 & Ministral-3  \\
\midrule
Numerical precision & BF16 & BF16 \\
LoRA rank & 128 & 128  \\
Learning rate & $10^{-5}$ & $2\times10^{-5}$  \\
Weight decay & - & 0.01 \\
Warmup updates (ToolQA, OfficeBench, ALFWorld, WebShop) & 13, 7, 39, 39 (linear) & - \\
Schedule after warmup & Constant & Constant  \\
Distillation temperature & 1.0 & 1.0  \\
Maximum generated tokens per action& 2,048 & 2,048  \\
\bottomrule
\end{tabular}

\end{table*}

\begin{table}[t]
\centering
\caption{Decoding parameters for the teacher (data collection) and the student (data collection+evaluation).}
\label{tab:eval_decoding}
\begin{tabular}{lcccc}
\toprule
Model & Temperature (teacher) & Top-$p$ (teacher) &Temperature (student) & Top-$p$ (student) \\
\midrule
Qwen3-8B & 0.7 & 0.95& 0.7 & 0.8 \\
Qwen3-14B & 0.7 & 0.95& 0.7 & 0.8 \\
Ministral-3 & 0.09 & 1.0& 0.08 & 1.0 \\
\bottomrule
\end{tabular}
\end{table}

\subsubsection{Trajectory Annotation Protocols}
\label{app:traj_analysis}

We use a two-stage annotation protocol. The first stage identifies exercises that satisfy the prerequisite for a given analysis, such as containing an issue that remains unresolved in the history. The second stage performs issue- or action-level annotation on the retained exercises. We primarily use Qwen3-235B-A22B as the annotator and use GPT-5.4 to adjudicate cases for which the primary annotator returns \texttt{Uncertain}.

\paragraph{Identifying exercises with unresolved issues.}
For each sampled exercise, we provide the complete interaction history to Qwen3-235B-A22B and ask it to identify all issues encountered during the trajectory and summarize how the agent responds to each issue. An issue may correspond, for example, to an incorrect intermediate decision, a failed operation, or a constraint revealed by environment feedback. Based on this analysis, the annotator determines whether at least one identified issue remains unresolved at the point of interest. It outputs one of three labels: \texttt{Yes}, \texttt{No}, or \texttt{Uncertain}, together with the identified issues and a brief summary of the attempted or completed resolution for each issue. Exercises labeled \texttt{Uncertain} are reassessed by GPT-5.4, which receives the original exercise together with the complete annotation produced by Qwen3-235B-A22B.

\paragraph{Annotating issue-specific agent behavior.}
After selecting exercises satisfying the prerequisite of the corresponding analysis, we perform more targeted annotations of the agent's behavior. Depending on the diagnostic, we use the following procedures:
\begin{itemize}
    \item \textbf{Issue recognition and corrective behavior.}
    We ask Qwen3-235B-A22B to determine whether the reasoning segment of the current action explicitly recognizes an issue present in the preceding history. If an issue is recognized, the annotator further determines whether the action (i) resolves the issue or (ii) takes an appropriate corrective step toward resolving it. We additionally provide the task-specific privileged information used in our distillation framework as reference information to help the annotator interpret the task requirements and the intended solution procedure. Cases labeled \texttt{Uncertain} are reassessed by GPT-5.4.

    \item \textbf{Quality of the proposed correction.}
    For annotations requiring a judgment about the quality of an action---for example, whether a proposed correction is executable and appropriate, or whether the action constitutes a valid self-correction---we use a solution-first annotation procedure. Before evaluating the agent's action, the annotator is first asked to independently derive an appropriate next step given the task history and available task information. It then evaluates the agent's proposed action relative to this independently derived solution. We found this decomposition to produce more reliable judgments than directly asking the annotator to assess action quality.
\end{itemize}
\subsection{Experiment results}
\subsubsection{Reverse KL}
\label{app:reverse_kl}
To compare with the main results in Table~\ref{tab:stage1_main_results} with forward KL, in Table~\ref{tab:stage1_main_results_reverse_kl}, we show the difference after applying reverse KL.

\begin{table*}[t]
\centering

\caption{
Comparison with results in Table~\ref{tab:stage1_main_results}. Arrows denote the value change.}
\label{tab:stage1_main_results_reverse_kl}
\resizebox{0.99\textwidth}{!}{%
\begin{tabular}{
l
ccc
ccc
ccc
cccc
}
\toprule
\multirow{2}{*}{\textbf{Model}}
& \multicolumn{3}{c}{\textbf{ToolQA}}
& \multicolumn{3}{c}{\textbf{OfficeBench}}
& \multicolumn{3}{c}{\textbf{ALFWorld}}
& \multicolumn{4}{c}{\textbf{WebShop}} \\
\cmidrule(lr){2-4}
\cmidrule(lr){5-7}
\cmidrule(lr){8-10}
\cmidrule(lr){11-14}

& \textbf{SR}
& \textbf{pass@3}
& \textbf{Action Count}
& \textbf{SR}
& \textbf{pass@3}
& \textbf{Action Count}
& \textbf{SR}
& \textbf{pass@3}
& \textbf{Action Count}
& \textbf{SR}
& \textbf{pass@3}
& \textbf{Task Score}
& \textbf{Action Count} \\
\midrule

Qwen3-8B
&& && & & & & & & && & \\
\quad + TT
&$\downarrow0.1$&$\downarrow0.4$ &$\downarrow0.2$&$\downarrow0.5$ &$\downarrow0.6$ &$\uparrow0.3$ &$\downarrow0.4$ & $\downarrow0.4$&  $\downarrow0.5$&$\downarrow0.6$&$\downarrow0.7$ &$\downarrow1.1$ &$\downarrow0.6$\\
\quad + ST
&$\downarrow0.1$&$\downarrow0.2$ &$\uparrow0.4$&$\downarrow0.2$ &$\downarrow0.4$ &$\uparrow0.2$ &$\uparrow0.2$ & $\downarrow0.2$&  $\downarrow0.1$&$\uparrow0.5$&$\downarrow0.3$ &$\downarrow0.3$ &$\uparrow0.4$\\
\quad + TS
&$\downarrow0.3$&$\downarrow0.1$ &$\downarrow0.4$&$\downarrow0.2$ &$\downarrow0.5$ &$\downarrow0.1$ &$\downarrow0.5$ & $\downarrow0.3$&  $\downarrow0.1$&$\downarrow0.6$&$\downarrow0.7$ &$\downarrow0.3$ &$\uparrow1.0$\\
\quad + SS
&$\uparrow0.2$&$\uparrow0.4$ &$\downarrow0.3$&$\downarrow0.1$ &$\downarrow0.6$ &$\uparrow0.2$ &$\uparrow0.2$ & $\downarrow0.3$&  $\downarrow0.1$&$\uparrow0.3$&$\downarrow0.4$ &$\uparrow0.2$ &$\downarrow0.2$\\

\midrule

Qwen3-14B
&& && & & & & & & && & \\
\quad + TT
&$\downarrow0.2$&$\downarrow0.4$ &$\uparrow0.3$&$\downarrow0.1$ &$\downarrow0.7$ &$\uparrow0.2$ &$\downarrow0.5$ & $\uparrow0.2$&  $\downarrow0.6$&$\downarrow0.2$&$\downarrow0.4$ &$\downarrow0.6$ &$\uparrow0.3$ \\

\quad + ST
&$\downarrow0.3$&$\downarrow0.5$ &$\uparrow0.1$&$\downarrow0.3$ &$\downarrow0.4$ &$\downarrow0.1$ &$\downarrow0.3$ & $\downarrow0.1$&  $\downarrow0.3$&$\downarrow0.5$&$\downarrow0.7$ &$\downarrow0.7$ &$\downarrow0.2$ \\
\quad + TS
&$\downarrow0.8$&$\downarrow0.6$ &$\downarrow0.2$&$\downarrow0.1$ &$\downarrow0.6$ &$\uparrow0.2$ &$\downarrow0.2$ & $\downarrow0.4$&  $\uparrow0.4$&$\downarrow0.1$&$\downarrow0.5$ &$\downarrow0.6$ &$\downarrow0.2$ \\
\quad + SS
&$\downarrow0.1$&$\uparrow0.1$ &$\downarrow0.3$&$\uparrow0.2$ &$\uparrow0.3$ &$\uparrow0.3$ &$\downarrow0.4$ & $\downarrow0.6$&  $\uparrow0.1$&$\downarrow0.2$&$\downarrow0.1$ &$\downarrow0.8$ &$\downarrow0.1$ \\

\midrule

Ministral-3
&& && & & & & & & && & \\
\quad + TT
&$\downarrow0.4$&$\downarrow0.6$ &$\uparrow0.2$&$\downarrow0.1$ &$\downarrow0.5$ &$\uparrow0.2$ &$\downarrow0.4$ & $\downarrow0.4$&  $\downarrow0.1$&$\downarrow0.2$&$\downarrow0.5$ &$\downarrow0.3$ &$\uparrow0.1$ \\
\quad + ST
&$\downarrow0.2$&$\downarrow0.3$ &$\uparrow0.2$&$\downarrow0.4$ &$\downarrow0.6$ &$\uparrow0.2$ &$\downarrow0.1$ & $\downarrow0.3$&  $\uparrow0.2$&$\downarrow0.1$&$\downarrow0.4$ &$\downarrow0.5$ &$\downarrow0.3$ \\
\quad + TS
&$\downarrow0.1$&$\downarrow0.5$ &$\uparrow0.1$&$\downarrow0.4$ &$\downarrow0.7$ &$\uparrow0.3$ &$\downarrow0.2$ & $\downarrow0.5$&  $\downarrow.1$&$\downarrow0.3$&$\downarrow0.8$ &$\downarrow0.5$ &$\uparrow0.1$ \\
\quad + SS
&$\downarrow0.2$&$\downarrow0.2$ &$\downarrow0.1$&$\uparrow0.1$ &$\uparrow0.4$ &$\uparrow0.2$ &$\downarrow0.2$ & $\downarrow0.5$&  $\uparrow0.2$&$\downarrow0.1$&$\downarrow0.3$ &$\downarrow0.3$ &$\downarrow0.2$ \\

\bottomrule
\end{tabular}
}
\end{table*}

\subsubsection{Stage 1: Analysis}
\label{app:stage1_mistake_analysis}
\begin{figure*}[t]
    \centering
    \includegraphics[width=0.5\textwidth]{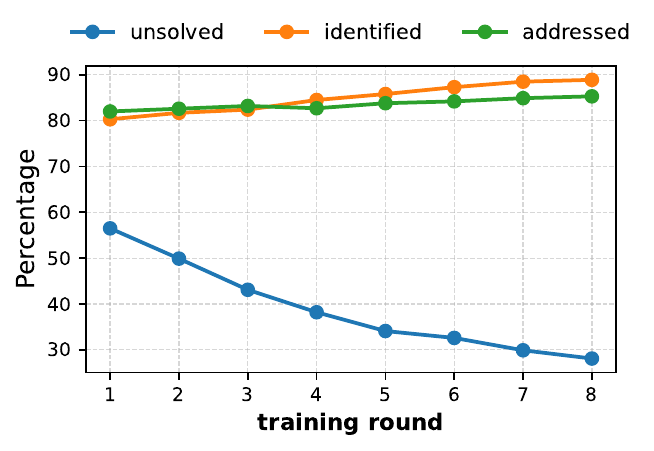}
    \caption{Evolution of corrective supervision across training rounds for Qwen3-14B on ALFWorld.}
    \label{fig:stage1_tt_vs_st}
\end{figure*}

\begin{figure*}[t]
    \centering
    \includegraphics[width=0.95\textwidth]{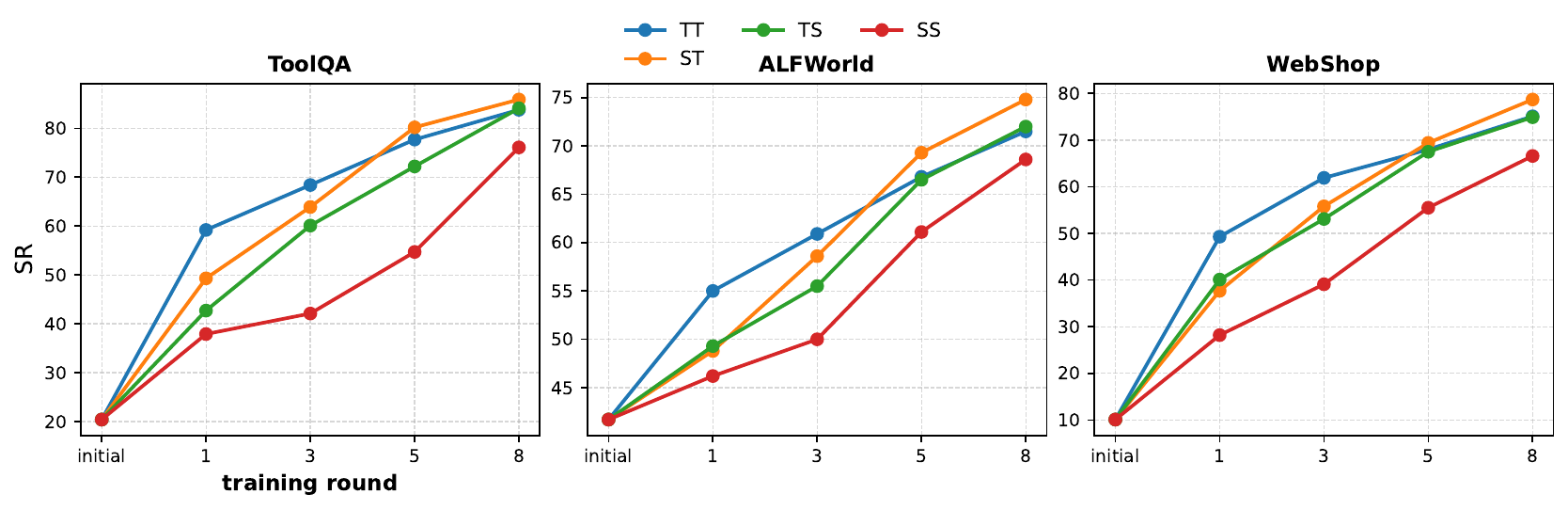}
    \caption{Validation success rate for Qwen3-14B on ToolQA, ALFWorld, and WebShop before training and after rounds 1, 3, 5, and 8.}
    \label{fig:stage1_validation}
\end{figure*}

\begin{figure*}[t]
    \centering
    \includegraphics[width=0.95\textwidth]{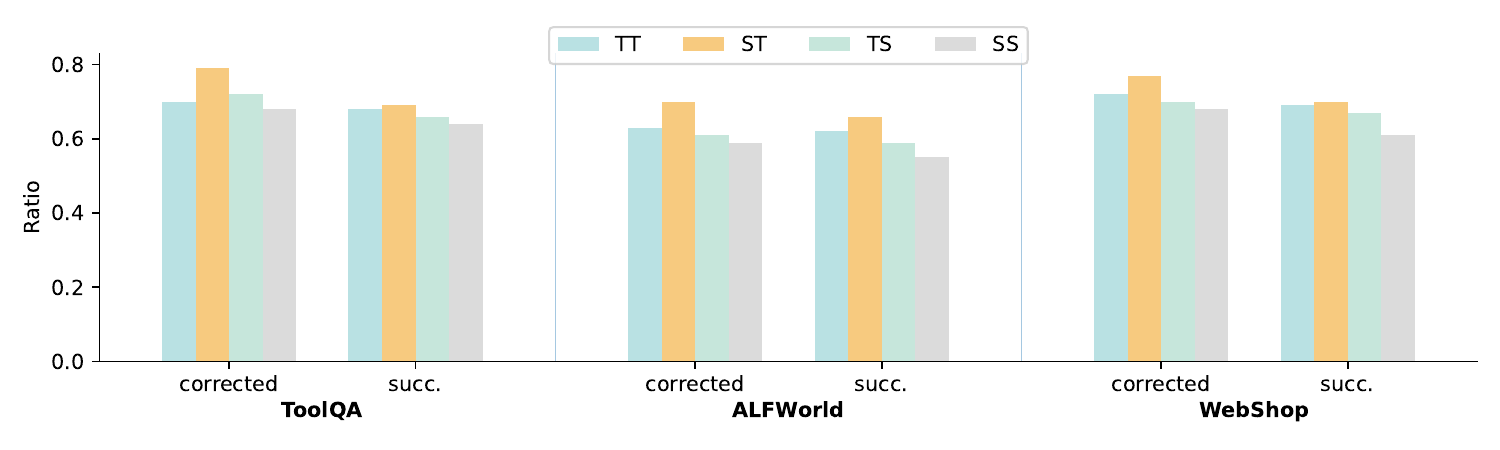}
    \caption{Recovery evaluation with Qwen3-14B on shared erroneous histories from initial-model failures. We evaluate 100 histories per benchmark. ``corrected'' measures correction of the existing issue, and ``succ.'' measures eventual task success.}
    \label{fig:stage1_recovery}
\end{figure*}

\begin{figure*}[t]
    \centering
    \includegraphics[width=0.95\textwidth]{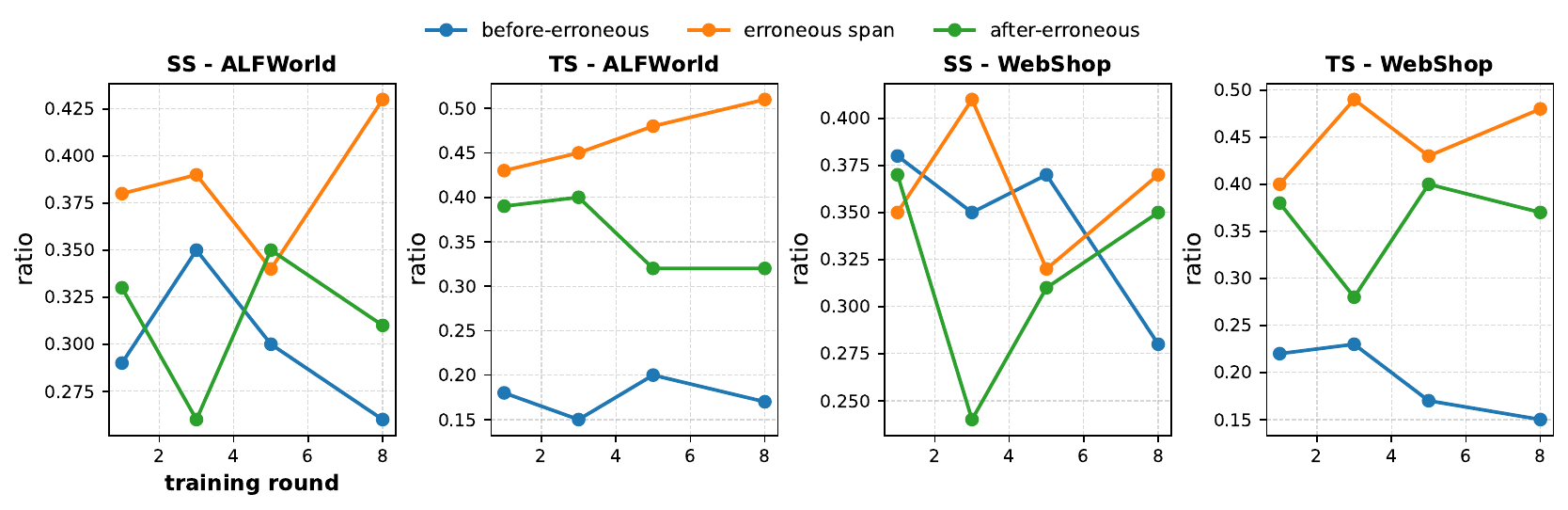}
    \caption{Regional teacher--student disagreement around the first annotated error in $SS$ and $TS$ training actions. At each selected round, we compute mean per-token KL before, within, and after the error span and normalize the three regional means by their sum.}
    \label{fig:stage1_ss_vs_ts}
\end{figure*}

Here we present results on trajectories and supervision analysis in Stage 1 in Table~\ref{tab:stage_1_tt_vs_st}, Table~\ref{tab:stage_1_mistake_type}, Table~\ref{tab:stage_1_ss_vs_st}, Figure~\ref{fig:stage1_tt_vs_st}, Figure~\ref{fig:stage1_validation}, Figure~~\ref{fig:stage1_ss_vs_ts} and Figure~\ref{fig:stage1_recovery}.

\begin{table}[t]
\begin{minipage}{0.9\textwidth}
\centering
\small
\caption{Corrective supervision in $TT$ and $ST$ training exercises. Unresolved measures the percentage of histories containing an uncorrected earlier error. Identified is conditional on an unresolved error; Addressed is conditional on identification and includes either resolution or an appropriate corrective step.}
\label{tab:stage_1_tt_vs_st}

\resizebox{\linewidth}{!}{%
\begin{tabular}{lcccc}
\toprule
Model & Configuration & Unresolved(\%) & Identified(\%) & Addressed(\%) \\
\midrule
\multirow{2}{*}{\textbf{Qwen3-8B}}
    & ST & 85.0&77.1 & 83.0\\ 
    & TT & 19.1& 90.2 &81.2 \\ 
\midrule
\multirow{2}{*}{\textbf{Qwen3-14B}}
    & ST & 82.7&80.7 & 84.5 \\
    & TT & 14.0& 89.3&86.0 \\ 
\midrule
\multirow{2}{*}{\textbf{Ministral-3}}
    & ST & 83.3 & 80.6&82.9 \\ 
    & TT & 15.7&88.1 & 82.9 \\ 
\bottomrule
\end{tabular}%
}
\end{minipage}
\end{table}

\begin{table*}[t]
\centering
\caption{Error profiles of trained students. Error categories are nonexclusive, so a trajectory may receive multiple labels.}
\label{tab:stage_1_mistake_type}
\resizebox{1.1\textwidth}{!}{\begin{tabular}{lcccccc}
\toprule
Model & Configuration &Repeated failure  & Premature completion & Rule inconsistency & Operational mistake & Planning mistake\\
\midrule
\multirow{4}{*}{\textbf{Qwen3-8B}} 
    & TT & 20\% & 7\% & \textbf{2}\% &\textbf{2}\% &11\%\\ 
    & ST & \textbf{9}\% & \textbf{5}\%& 7\%& \textbf{2}\%&\textbf{7}\%\\ 
    & TS & 17\% & 7\%& 7\%& 5\%&\textbf{7}\%\\ 
    & SS & 22\% &24\% &7\% & 12\% &20\%\\  
\midrule
\multirow{4}{*}{\textbf{Qwen3-14B}}
    & TT &14\% &7\% &\textbf{1}\% &2\% &9\%\\ 
    & ST & \textbf{7}\%& \textbf{2}\%& 6\%&2\% &\textbf{5}\%\\ 
    & TS & 15\%& 7\%&5\% & \textbf{1}\%&7\%\\ 
    & SS &20\% & 17\%& 9\%&11\% &19\%\\  
\midrule
\multirow{4}{*}{\textbf{Ministral-3}}
    & TT & 17\%&9\%&\textbf{4}\% & \textbf{1}\%&10\%\\ 
    & ST & \textbf{7}\%&\textbf{4}\%& \textbf{4}\%&2\% &\textbf{5}\%\\ 
    & TS & 15\%&9\%& 5\%&\textbf{1}\% &12\%\\ 
    & SS &22\% &20\%&9\% &9\% &22\%\\  
\bottomrule
\end{tabular}}
\end{table*}

\begin{table}
\begin{minipage}{0.5\textwidth}
\centering

\caption{Teacher-continuation quality at matched student histories with unresolved errors. The teacher generates either the entire current action or a continuation after the retained student reasoning, optionally followed by the first half of the student's operation.}
\label{tab:stage_1_ss_vs_st}

\resizebox{\linewidth}{!}{%
\begin{tabular}{lccc}
\toprule
Model&Retained student prefix & Executability (\%) & Appropriateness (\%)  \\
\midrule
\multirow{2}{*}{\textbf{Qwen3-8B}}
    & None ($ST$) & 92.0 & 80.2\\ 
    & Reasoning &81.0 & 70.7  \\ 
& Reasoning + half operation&73.2 & 56.2  \\ 
\midrule
\multirow{2}{*}{\textbf{Qwen3-14B}}
    & None ($ST$) & 94.7 & 83.8\\ 
    & Reasoning & 85.2 & 76.0  \\ 
& Reasoning + half operation & 76.0& 60.2\\ 
\bottomrule
\end{tabular}
}
\end{minipage}
\end{table}

\subsubsection{TT example}
\label{app:stage1_tt_example}

\subsubsection{Stage 2 Main results (cont.)}
\label{app:stage2_main}
\begin{figure*}[t]
    \centering
    \includegraphics[width=0.95\textwidth]{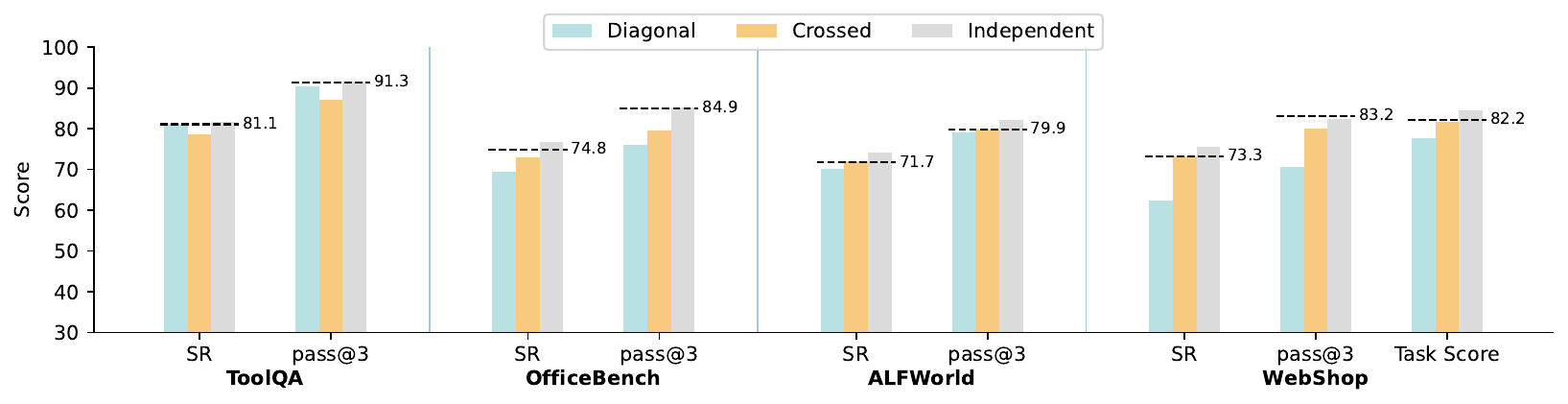}
    \caption{Pairing at fixed marginal on-policiness with Qwen3-8B. Dashed lines indicate standalone $ST$ performance from Stage~1.}
    \label{fig:stage2_main_qwen3-8b}
\end{figure*}

\begin{figure*}[t]
    \centering
    \includegraphics[width=0.95\textwidth]{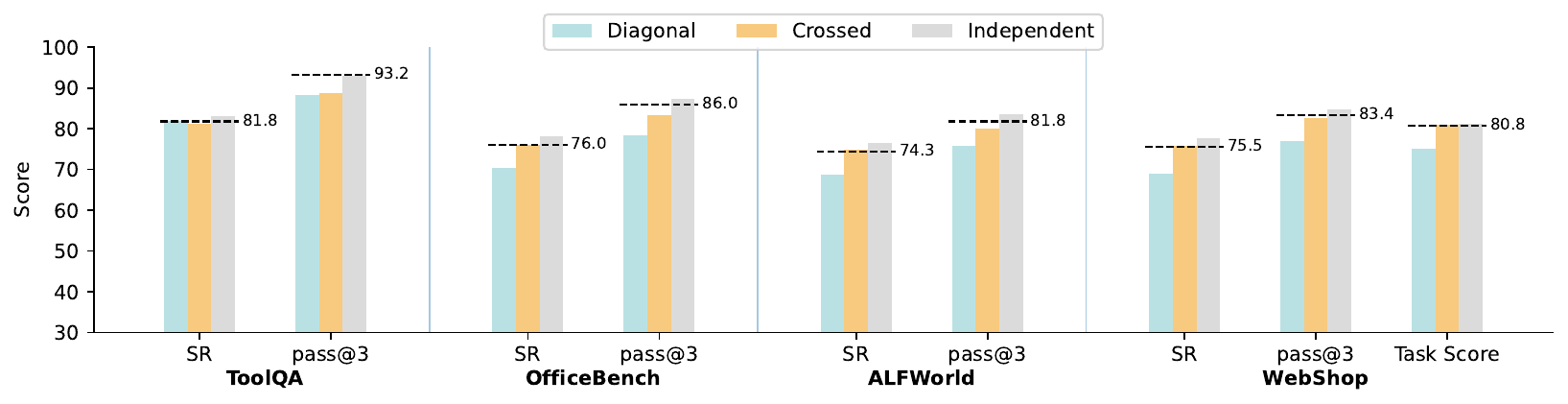}
    \caption{Pairing at fixed marginal on-policiness with Ministra-3. Dashed lines indicate standalone $ST$ performance from Stage~1.}
    \label{fig:stage2_main_ministral}
\end{figure*}

Table~\ref{fig:stage2_main_qwen3-8b} and Table~\ref{fig:stage2_main_ministral} show results on Qwen3-8B and Ministral-3, respectively.

\subsubsection{Stage 2: TT and SS contributions decomposition}

\begin{figure*}[t]
    \centering
    \includegraphics[width=0.95\textwidth]{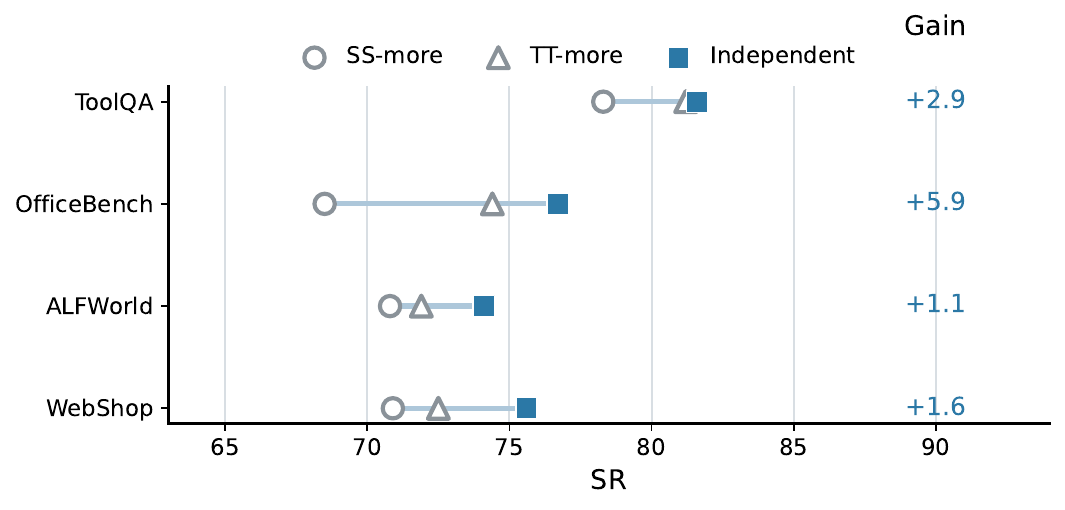}
    \caption{Retaining both diagonal configurations improves success rate with Qwen3-8B. Holding $ST$ and $TS$ at 0.25 each, the remaining allocation is assigned to $TT$ only (TT-more), $SS$ only (SS-more), or equally to both (Independent)}
    \label{fig:stage2_tt_vs_ss_qwen3-8b}
\end{figure*}

\begin{figure*}[t]
    \centering
    \includegraphics[width=0.95\textwidth]{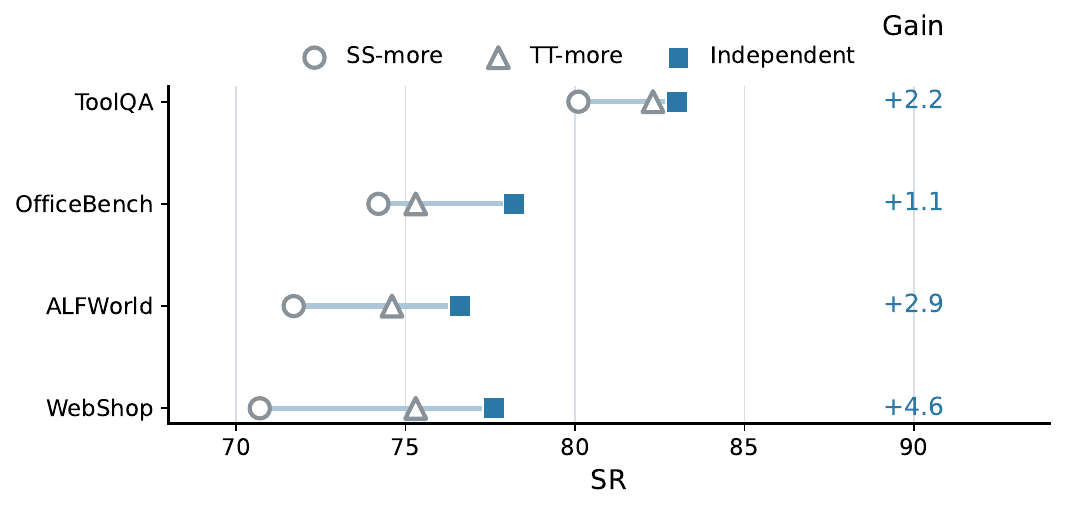}
    \caption{Retaining both diagonal configurations improves success rate with Ministral-3. Holding $ST$ and $TS$ at 0.25 each, the remaining allocation is assigned to $TT$ only (TT-more), $SS$ only (SS-more), or equally to both (Independent)}
    \label{fig:stage2_tt_vs_ss_ministral}
\end{figure*}

\subsubsection{Stage 2: ST vs TS}

\begin{figure*}[t]
    \centering
    \includegraphics[width=0.95\textwidth]{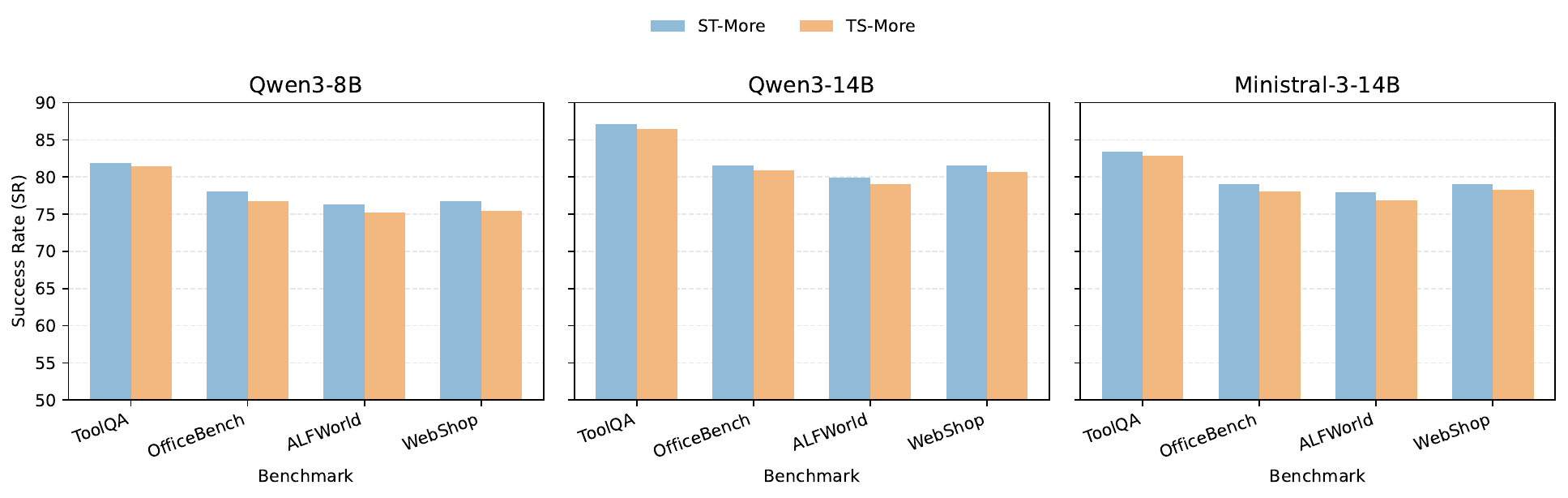}
    \caption{Comparing the two hybrid directions. ST-more uses weights $(0.125,0.5,0.25,0.125)$, while TS-more uses $(0.125,0.25,0.5,0.125)$, ordered as $(TT,ST,TS,SS)$.}
    \label{fig:stage2_st_vs_ts}
\end{figure*}

\subsubsection{Stage 2: Hybrid configuration}
\begin{table}
\begin{minipage}{0.5\textwidth}
\centering

\caption{Success-rate gains, in percentage points, from replacing the Independent mixture $(0.25,0.25,0.25,0.25)$ with the hybrid-heavy mixture $(0.125,0.375,0.375,0.125)$.}
\label{tab:stage_2_more_st_ts}

\resizebox{\linewidth}{!}{%
\begin{tabular}{lcccc}
\toprule
Model&ToolQA & OfficeBench & ALFWorld &  WebShop\\
\midrule
Qwen3-8B&0.1& 0.7&1.9  & 0.7\\
Qwen3-14B&0.3& 1.4& 1.6 &0.5 \\
Ministral-3& 0.3&0.6 & 0.7 & 0.9\\
\bottomrule
\end{tabular}
}
\end{minipage}
\end{table}

\end{document}